\documentclass[letterpaper, 10 pt, conference]{ieeeconf} 
\IEEEoverridecommandlockouts
\usepackage{cite}
\usepackage{amsmath,amssymb,amsfonts}
\usepackage{algorithmic}
\usepackage{graphicx}
\usepackage{textcomp}
\usepackage{xcolor}
\usepackage{times}
\usepackage{comment}
\usepackage{epsfig}
\usepackage{graphicx}
\usepackage{amsmath}
\usepackage{amssymb}
\usepackage{graphicx}
\usepackage{url}
\usepackage{array}
\usepackage{multirow}
\usepackage{subcaption}
\usepackage{color}
\usepackage{resizegather}
\usepackage{makecell}
\usepackage{textcomp}
\usepackage{resizegather}
\usepackage{cite}
\usepackage{footnote}
\def\BibTeX{{\rm B\kern-.05em{\sc i\kern-.025em b}\kern-.08em
    T\kern-.1667em\lower.7ex\hbox{E}\kern-.125emX}}
\begin{document}

\title{CSGAN: Cyclic-Synthesized Generative Adversarial Networks for Image-to-Image Transformation
%
}
\author{Kishan Babu Kancharagunta and Shiv Ram Dubey
\thanks{K.K. Babu and S.R. Dubey  are with Computer Vision Group, Indian Institute of Information Technology, Sri City, Andhra Pradesh - 517646, India. 
{\tt\small email: \{kishanbabu.k, srdubey\}@iiits.in}}%
}

\maketitle

\begin{abstract}
The primary motivation of Image-to-Image Transformation is to convert an image of one domain to another domain. Most of the research has been focused on the task of image transformation for a set of pre-defined domains. Very few works are reported that actually developed a common framework for image-to-image transformation for different domains. With the introduction of Generative Adversarial Networks (GANs) as a general framework for the image generation problem, there is a tremendous growth in the area of image-to-image transformation. Most of the research focuses over the suitable objective function for image-to-image transformation. In this paper, we propose a new Cyclic-Synthesized Generative Adversarial Networks (CSGAN) for image-to-image transformation. The proposed CSGAN uses a new objective function (loss) called Cyclic-Synthesized  Loss (CS) between the synthesized image of one domain and cycled image of another domain.  The performance of the proposed CSGAN is evaluated on two benchmark image-to-image transformation datasets, including CUHK Face dataset and CMP Facades dataset. The results are computed using the widely used evaluation metrics such as MSE, SSIM, PSNR, and LPIPS. The experimental results of the proposed CSGAN approach are compared with the latest state-of-the-art approaches such as GAN, Pix2Pix, DualGAN, CycleGAN and PS2GAN. The proposed CSGAN technique outperforms all the methods over CUHK dataset and exhibits the promising and comparable performance over Facades dataset in terms of both qualitative and quantitative measures. The code is available at \textcolor{blue}{https://github.com/KishanKancharagunta/CSGAN}
\end{abstract}


\section{Introduction}
\label{introduction}
\begin{figure}[!t]
\begin{center}
\includegraphics[width=\linewidth , height=6 cm]{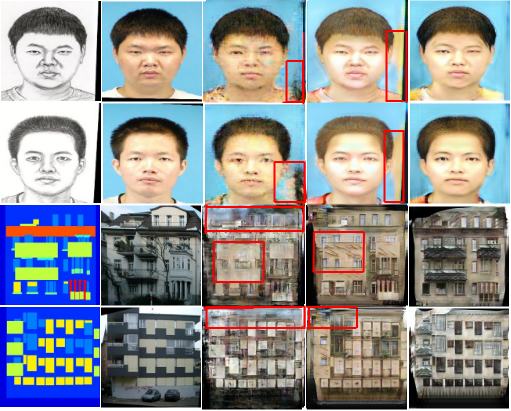}
 \setlength{\belowcaptionskip}{-20pt} 
\caption{Sample results from CUHK dataset \cite{chuk} in $1^{st}$ and $2^{nd}$ rows and FACADES dataset \cite{facades} in $3^{rd}$ and $4^{th}$ rows. The $1^{st}$  and $2^{nd}$ columns show the input images and ground truth images, respectively. The $3^{rd}$, $4^{th}$, and $5^{th}$ columns represent the transformed images using DualGAN \cite{dualGAN}, CycleGAN \cite{CyclicGAN}, and proposed CSGAN, respectively. Note that the artifacts in DualGAN and CycleGAN are marked with red color rectangles in $2^{nd}$ and $3^{rd}$ columns, respectively.}
\label{fig:qualititative}
\end{center}
\end{figure}

Recent advancements in image-to-image transformation problems, in which the image from one domain is transformed to the corresponding image of another domain. The domain specific problem has its applications in the fields of image processing, computer graphics and computer vision 
that includes image colorization 
\cite{DC,CIC}
image super-resolution 
\cite{DLISR,SISRDL}
image segmentation 
\cite{SISDPN,SASIS}
image style transfer \cite{ISTCNN,PLRTSTSR}
and 
face photo-sketch synthesis 
\cite{FSSSRGS,fccn}.
In this paper, a cyclic-synthesized generative adversarial network (CSGAN) is proposed for the image-to-image transformation. Fig. \ref{fig:qualititative} highlights the improved performance of the CSGAN method for sketch to face synthesis compared to the latest state-of-the-art methods.  

Traditionally, the above mentioned image-to-image transformation problems are handled by different transformation mechanisms \cite{chuk}, \cite{NLAID}, as per the need. 
 Even, some of the image-to-image transformation problems are dealt with other strategies such as classification, regression, etc.  
 A multi-scale Markov random fields (MRF) based face photo-sketch synthesis model is proposed to transfer the face sketch into a photo and vice-versa \cite{chuk}. An image quilting method for texture synthesis is presented by \cite{chuk} as an image transfer problem that uses patch based image stitching. A non local means method is proposed for image denoising based on an average of all pixel values in the image \cite{NLAID}. 
\begin{figure*}[t]
\begin{center}
\includegraphics[width=\linewidth, height= 10cm]{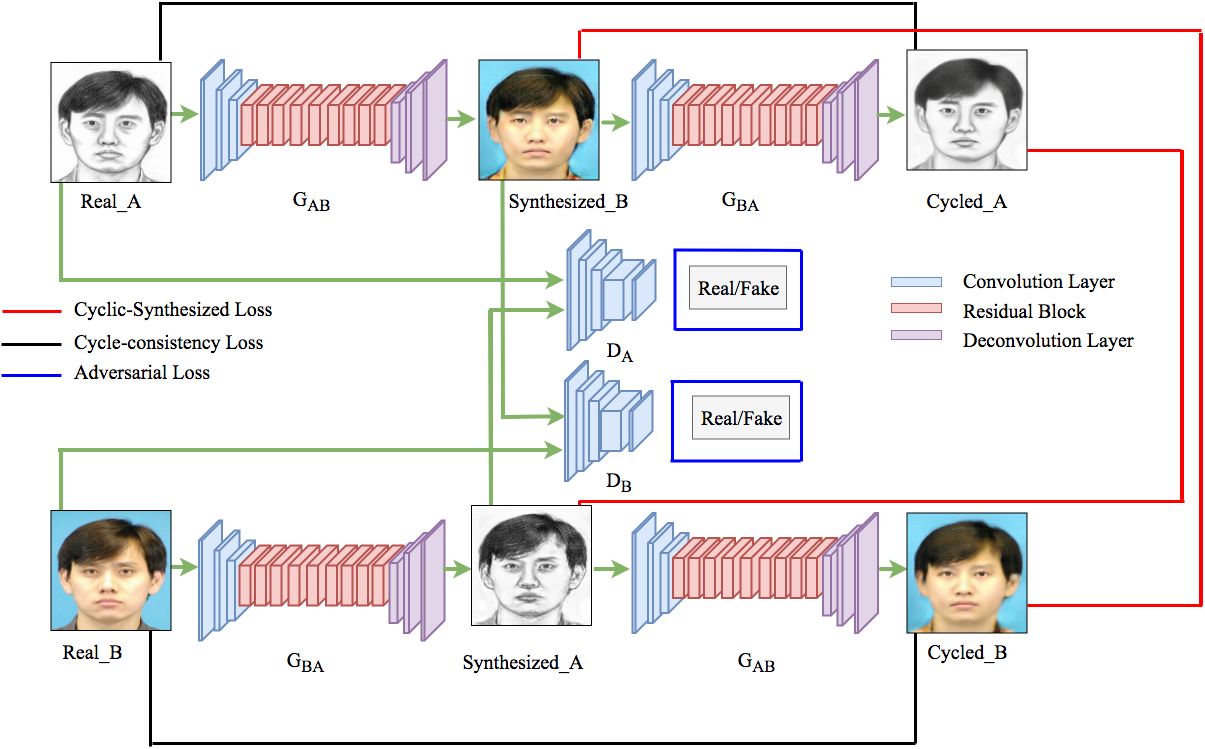}
 \setlength{\belowcaptionskip}{-18pt} 
\caption{Network architecture of the proposed CSGAN for image-to-image transformation. The cyclic-synthesized loss is proposed in this paper to utilize the relation between synthesized and cycled images in both the image domains. Thus, in addition to the adversarial loss and cycle-consistency loss, we have used cyclic-synthesized loss also to train the network. The adversarial loss is represented in the blue color rectangles which is calculated between 1) the generator $G_{AB}$ and the discriminator $D_B$, and 2) the generator $G_{BA}$ and the discriminator $D_A$. The cycle-consistency loss is shown in the black color and is calculated as $L_1$ loss between the real and cycled images. The cyclic-synthesized loss is shown in the red color and is calculated as $L_1$ loss between the synthesized and cycled images.}
\label{CSGAN}
\end{center}
\end{figure*}
 
 Later with the development of Deep Learning, Convolutional Neural Networks (CNNs) became very popular and widely used for different computer vision problems 
like object recognition 
 \cite{ICDCN}, 
object localization 
 \cite{EOLCNN}, 
 human action prediction 
 \cite{3DCNNHAR} and
 medical image analysis 
 \cite{CNNMIA}. 
The CNN based methods \cite{DCLR}
and \cite{ISTCNN} for image-to-image transformation automatically learn the transformation function.
 It depends on the network architecture in the training phase with the given loss function. For example, \cite{DCLR} implemented a CNN based network for image colorization that uses Euclidean distance (i.e., $L_2$) as a loss function. A colorful image colorization network is designed by \cite{CIC} which uses the multinomial cross entropy loss and gives better results compared to the \cite{DCLR} because of the averaging tendency of the Euclidean distance. \cite{ISTCNN} used combination of squared error and mean square error as the loss function for the image style transfer. 
Even though, the CNN based methods learn the transformation automatically, its performance depends on the selection of the loss functions that best suit for particular domain transformation.

Goodfellow et al. proposed the Generative Adversarial Network (GAN) for image generation in a given dataset \cite{gan}. It uses two networks, namely generator to generate the new samples and discriminator to distinguish between the generated and real samples. 
 The competitive training of generator and discriminator is done such that the generator learns how to generate more realistic fake image, whereas the discriminator learns how to distinguish between the generated high quality fake image and real image. 
Generative Adversarial Networks (GANs) were initially proposed to generate models that are nearer to the training samples based on the given noise distribution. Later on the GANs were also used for different applications 
like image blending \cite{TRHRIB}, semantic segmentation \cite{SSUAN}, single image super-resolution \cite{PRSISRGAN} and image inpainting \cite{FLIP}, etc. These methods were introduced for particular applications, thus the generalized framework is still missing.

Isola et al. explored the Pix2Pix as the $1^{st}$ common framework to work on image-to-image translation using Conditional GANs (cGANs) for paired image datasets \cite{cGAN}. 
Wang et al. proposed a perceptual adversarial networks (PANs) for solving image-to-image transformation problems. The PANs combines the perceptual adversarial loss with the generative adversarial loss to solve this problem \cite{PAN}. 
 The perceptual adversarial loss of PAN enforces the network to learn the similarity between image pairs more semantically. 
Zhu et al. presented a framework with inverse mapping function for unsupervised data using Cycle-Consistent Adversarial Network (CycleGAN) \cite{CyclicGAN}. 
 The CyclicGAN transfer the images in two domains in both ways, i.e., in froward as well as in backward direction.
Yi et al. developed a framework for image-to-image translation in an unsupervised setting using Dual-GAN mechanism \cite{dualGAN}. Wang et al. proposed a framework for photo-sketch synthesis involving multi-adversarial networks (PS2MAN) \cite{ps2-man}. 

 In spite of the above mentioned recent developments, there are still gaps in terms of network architecture of the generator and discriminator, restrain on the size of the datasets and choice of the objective functions. 
 The above mentioned network architectures mainly differ in terms of the loss functions used for the training. 
Most of the recent works included the Adversarial loss calculated between generators and discriminators, the Cycle-consistency loss calculated between the Real\_Images and Cycled\_Images, and the Synthesized loss calculated between the Real\_Images and Synthesized\_Images. 
All these losses are used to minimize the gap between real and generated images. 
Even after considering all these losses, we still find the scope to minimize the loss between the Synthesized\_Images and Cycled\_Images. As per the best of our knowledge, no existing network utilizes the loss between the Synthesized\_Images and Cycled\_Images. 
 Synthesized\_Images are generated from the generators by giving Real\_Images as the input and the same generators are used to generate the Cycled\_Images by taking the Synthesized\_Images as the input. 
In this paper, we propose a new loss function called as the Cyclic\_Synthesized loss, which is first of its kind.

The contributions of this paper are mainly three-fold:
\begin{itemize}
    \item We proposed a new loss function Cyclic-Synthesized Loss (CS Loss) that increases the quality of the results produced with reduced artifacts.
    \item We proposed the CSGAN architecture based on the CS Loss for image-to-image transformation.
    \item We evaluated our method over two benchmark datasets with baseline image quality assessment metrics and our method shows the better/comparable performance as compared to the state-of-the-art methods.
\end{itemize}

\section{Proposed CSGAN Architecture}
In this section, first, we describe the problem formulation, followed by the depiction of the proposed method and its objective function. Later, the generators and discriminators architecture are discussed in detail.

For a given dataset $X \in \{(A_i),(B_i)\}_{i=1}^{n}$ which consists of the $n$ number of the paired images of two different domains $A$ and $B$, the goal of our method is to train two transformation functions, i.e., $G_{AB}:A \rightarrow B$ and $G_{BA}:B \rightarrow A$. The $G_{AB}$ is a generator that takes a Real\_Image $(R_A)$ from domain $A$ as the input and tries to transform it into a Synthesized\_Image $(Syn_B)$ of domain $B$. The $G_{BA}$ is another generator that takes a Real\_Image $(R_B)$ in domain $B$ as the input and tries to translate it into a Synthesized\_Image $(Syn_A)$ of domain $A$. In addition to the above two generators, the proposed method consists of two discriminators $D_A$ and $D_B$ to distinguish between $R_A$ and $Syn_A$ in domain $A$ and $R_B$ and $Syn_B$ in domain $B$, respectively. The CycleGAN also used these two discriminators \cite{CyclicGAN}. The real image of domain $A$, $R_A$ and the real image of domain $B$, $R_B$ are transformed into the synthesized image in domain $B$, $Syn_B$ and synthesized image in domain $A$, $Syn_A$, respectively as, 
\begin{equation}
\label{fake_imagesB}
Syn_B=G_{AB}(R_A)
\end{equation}
\begin{equation}
\label{fake_imagesA}
Syn_A=G_{BA}(R_B)
\end{equation}

The synthesized images (i.e., $Syn_B$ and $Syn_A$) are again transformed into cycled images in another domain (i.e., $Cyc_A$ and $Cyc_B$), respectively as,
\begin{equation}
\label{cycled_imagesA}
Cyc_A=G_{BA}(Syn_B)=G_{BA}((G_{AB}(R_A))
\end{equation}
\begin{equation}
\label{cycled_imagesB}
Cyc_B=G_{AB}(Syn_A)=G_{AB}(G_{BA}(R_B))
\end{equation}

The overall work of the proposed CSGAN method as shown in Fig. \ref{CSGAN}
 is to transform image $R_A$ from domain $A$ to $B$ 
by giving it to the generator network $G_{AB}$, results in the synthesized image $Syn_B$. The Synthesized image $Syn_B$ from domain $B$
 and again transformed into the original domain $A$.
by giving it to the generator network $G_{BA}$, results in the cycled image $Cyc_A$. 
 In the same way the real image $R_B$ from domain $B$ is first transformed into the domain $A$ 
 as the synthesized image $Syn_A$
 and then transformed back into the domain $B$. 
 as the cycled image $Cyc_B$ by using the generator networks $G_{BA}$ and $G_{AB}$, respectively. The discriminator network $D_A$ is used to distinguish between the real image $R_A$ and synthesized image $Syn_A$. In the same way, the discriminator network $D_B$ is used to distinguish between the real image $R_B$ and synthesized image $Syn_B$. 
To generate the synthesized images nearest to the real images, the loss between them is to be minimized. This signifies the need for efficient loss function.

\subsection{Proposed Cyclic-Synthesized Loss }\label{CS_Loss}
In this paper, the Cyclic-Synthesized loss is proposed to reduce the above mentioned artifacts. 
The generator network $G_{BA}$ used to generate the Synthesized\_Image $Syn_A$ from the Real\_Image $R_B$ is also used to generate the Cycled\_Image $Cyc_A$ from the Synthesized\_Image $Syn_B$. In a similar way, the generator network $G_{AB}$ used to generate the Synthesized\_Image $Syn_B$ from the Real\_Image $R_A$ is also used to generate the Cycled\_Image $Cyc_B$ from the Synthesized\_Image $Syn_A$ as shown in Fig. \ref{CSGAN}. The distance between the Synthesized\_Image and the Cycled\_Image should be low as both are generated from the same generator. By this, the proposed Cyclic-Synthesized loss is calculated as $L_1$ loss between the Synthesized\_Image ($Syn_A$) and the Cycled\_Image ($Cyc_A$) in domain $A$ and the Synthesized\_Image ($Syn_B$) and the Cycled\_Image ($Cyc_B$) in domain $B$. The Cyclic-Synthesized loss is defined as follows,
\begin{equation}
\label{LCSA}
\mathcal{L}_{CS_A}=\left \| Syn_A-Cyc_A \right \|_1=\left \| G_{BA}(R_B)-G_{BA}(G_{AB}(R_A))\right\|_1
\end{equation}
\begin{equation}
\label{LCSB}
\mathcal{L}_{CS_B}=\left \| Syn_B-Cyc_B \right \|_1=\left \| G_{AB}(R_A)-G_{AB}(G_{BA}(R_B))\right\|_1
\end{equation}
where the $\mathcal{L}_{CS_A}$ is the Cyclic-Synthesized loss in domain $A$ (i.e., between $Syn_A$ and $Cyc_A$) and $\mathcal{L}_{CS_B}$ is the Cyclic-Synthesized loss in domain $B$ (i.e., between $Syn_B$ and $Cyc_B$).

\subsection{CSGAN Objective Function}
The objective function ($\mathcal{L}$) for the proposed CSGAN method combines the proposed Cyclic-Synthesized loss with existing Adversarial loss and Cycle-consistency loss as follows,
\begin{equation}
\label{LFINAL}
\begin{split}
\mathcal{L}_(G_{AB}, G_{BA}, D_A,
D_B)=\mathcal{L}_{LSGAN_A}+\mathcal{L}_{LSGAN_B} \\ +\lambda_A\mathcal{L}_{cyc_A} +\lambda_B{\mathcal{L}_{cyc_B}}+\mu_A\mathcal{L}_{CS_A}+\mu_B\mathcal{L}_{CS_B}.
\end{split}
\end{equation}
\label{LCYCA}

where $\mathcal{L}_{CS_A}$ and $\mathcal{L}_{CS_B}$ are the proposed Cyclic-Synthesized loss explained in subsection \ref{CS_Loss}; $\mathcal{L}_{LSGAN_A}$, $\mathcal{L}_{LSGAN_B}$ are the adversarial loss and $\mathcal{L}_{cyc_A}$, $\mathcal{L}_{cyc_B}$ are the Cycle-consistency loss proposed in CycleGAN \cite{CyclicGAN}. The adversarial loss and the Cycle-consistency loss are described in detail in following sub-sections:

\subsubsection{Adversarial Loss}
The generator networks $G_{AB}:A \rightarrow B$ and $G_{BA}: B \rightarrow A$ used in the proposed model are trained using the adversarial loss that comes from the discriminator against the generator network over a common objective function similar to the adversarial loss of original GAN \cite{gan}. The Generator network generates an image that looks similar to the original image, whereas the Discriminator distinguishes between the real and generated images. In this way both the Generator and discriminator networks are trained simultaneously by eliminating the problem of generating blurred images when $L_1$ or $L_2$ loss functions are used \cite{cGAN}. Similar to the CycleGAN \cite{CyclicGAN}, the least square loss introduced in \cite{LSGAN}, is used in the proposed method as the Adversarial loss. The least square loss stabilizes the training procedure to generate the high quality results. The adversarial loss between the Generator network $G_{AB}$ and the Discriminator network $D_B$ is computed as follows, 
\begin{equation}
\begin{split}
\label{LSGANB}
\mathcal{L}_{LSGAN_B}(G_{AB}, D_B, A, B) &= \mathbb{E}_{B \sim P_{data}(B)} [(D_B(R_B)-1)^2] \\ +\mathbb{E}_{A \sim P_{data}(A)} [D_B(G_{AB}(R_A))^2].
\end{split}
\end{equation}
Similarly, the adversarial loss between the Generator network $G_{BA}$ and the Discriminator network $D_A$ is computed as follows,
\begin{equation}
\begin{split}
\label{LSGANA}
\mathcal{L}_{LSGAN_A}(G_{BA}, D_A, B, A) &= \mathbb{E}_{A \sim P_{data}(A)} [(D_A(R_A)-1)^2] \\ 
+\mathbb{E}_{B \sim P_{data}(B)} [D_A(G_{BA}(R_B))^2].
\end{split}
\end{equation}
where $G_{AB} $ and $G_{BA}$ are used to transform the images from $A\rightarrow B$ and $B\rightarrow A$, respectively, and $D_A$ and $D_B$ are used to distinguish between the original and transformed images in the domains of $A$ and $B$, respectively. The Adversarial loss works as a good learned transformation function that can learn the distributions from the input images in training and generate similar looking images in testing. Even though Adversarial loss removes the problem of blurred images, still it produces the artifacts in the images due to the lack of the sufficient goodness measure.

\subsubsection{Cycle-consistency Loss}
The Cycle-consistency loss as discussed in \cite{CyclicGAN} is also used in the objective function of the proposed method. It is calculated as $L_1$ loss between the Real\_Image ($R_A$) and the Cycled\_Image ($Cyc_A$) in domain $A$ and the Real\_Image ($R_B$) and the Cycled\_Image ($Cyc_B$) in domain $B$. The Cycle-consistency loss is defined as follows,
\begin{equation}
\label{LCYCA}
\mathcal{L}_{cyc_A}=\left \| R_A-Cyc_A \right \|_1=\left \| R_A-G_{BA}(G_{AB}(R_A))\right\|_1
\end{equation}
\begin{equation}
\label{LCYCB}
\mathcal{L}_{cyc_B}=\left \| R_B-Cyc_B \right \|_1=\left \| R_B-G_{AB}(G_{BA}(R_B))\right\|_1
\end{equation}
where the $\mathcal{L}_{cyc_A}$ is the Cycle-consistency loss in domain $A$ (i.e., between $R_A$ and $Cyc_A$) and $\mathcal{L}_{cyc_B}$ is the Cycle-consistency loss in domain $B$ (i.e., between $R_B$ and $Cyc_B$).

We used $L_1$ distance instead of $L_2$ distance as the $L_2$ distance produces more blurred results when compared to the $L_1$ distance. The two losses $\mathcal{L}_{cyc_A}$ and $\mathcal{L}_{cyc_B}$ act as the forward and backward consistencies and introduce the constraints to reduce the space of possible mapping functions. Due to the large size of networks and with more mapping functions, the two losses serve the purpose of regularization while learning network parameters.

The above mentioned losses, i.e., Adversarial loss and Cycle-consistency loss used in CycleGAN \cite{CyclicGAN} and DualGAN \cite{dualGAN} produced good quality images. However, there is a  need to minimize the artifacts produced as shown in Figure \ref{fig:qualititative} for which we propose the Cyclic-Synthesized loss in this paper.

\subsection{Network Architectures}
In this paper, the Generator and Discriminator architectures are adapted from \cite{CyclicGAN}. The Generator network, as shown in Table\ref{Generator Network} 
consists of $3$ Convolutional Layers followed by $9$ residual blocks and $3$ Deconvolutional Layers, is basically adapted from \cite{PLRTSTSR}. 
 A brief description about the Generator and Discriminator networks are presented in this subsection.
The used Discriminator network in the proposed method is a $70 \times 70$ PatchGAN taken from \cite{cGAN}. This network consists of the $4$ Convolutional Layers, each one is a sequence of Convolution-InstanceNorm-LeakyReLU, followed by $1$ Convolutional Layer to produce a $1$ dimensional output.as shown in Table \ref{Discriminator Network}
In the $1^{st}$ convolutional layer, we do not use any normalization.

 \begin{table}[!t]
 \caption{Generator Network Architecture. The $r\_p$, $s$ and $p$ denote the size of reflection padding, stride, and padding, respectively.}
 \label{Generator Network}
 \begin{center}
 \scalebox{1}{
 \begin{tabular}{@{}l l}
 \hline 
 Input: & Image ($256$x$256$) \\
 \hline
 \hline
 [layer $1$]& r\_p=$3$; Conv2d $(7,7,64)$, s=$1$; ReLU;\\
 \hline
 [layer $2$]& Conv2d$(3,3,128)$, s=$2$, p=$1$; InstanceNorm; ReLU;\\
 \hline
 [layer $3$]& Conv2d $(3,3,256)$, s=$2$, p=$1$; InstanceNorm; ReLU;\\
 \hline
 [layer $4$] & \makecell{r\_p=$1$; Conv2d $(3,3,256)$, s=$1$; InstanceNorm; ReLU;\\
                      r\_p=$1$; Conv2d $(3,3,256)$, s=$1$; InstanceNorm;} \\
 \hline
 [layer $5$] & \makecell{r\_p=$1$; Conv2d $(3,3,256)$, s=$1$; InstanceNorm; ReLU;\\
                      r\_p=$1$; Conv2d $(3,3,256)$, s=$1$; InstanceNorm;} \\
 \hline
 [layer $6$] & \makecell{r\_p=$1$; Conv2d $(3,3,256)$, s=$1$; InstanceNorm; ReLU;\\
                      r\_p=$1$; Conv2d $(3,3,256)$, s=$1$; InstanceNorm;} \\
 \hline
 [layer $7$] & \makecell{r\_p=$1$; Conv2d $(3,3,256)$, s=$1$; InstanceNorm; ReLU;\\
                      r\_p=$1$; Conv2d $(3,3,256)$, s=$1$; InstanceNorm;} \\
 \hline
 [layer $8$] & \makecell{r\_p=$1$; Conv2d $(3,3,256)$, s=$1$; InstanceNorm; ReLU;\\
                      r\_p=$1$; Conv2d $(3,3,256)$, s=$1$; InstanceNorm;} \\ 
                      \hline
 [layer $9$] & \makecell{r\_p=$1$; Conv2d $(3,3,256)$, s=$1$; InstanceNorm; ReLU;\\
                      r\_p=$3$; Conv2d $(3,3,256)$, s=$1$; InstanceNorm;} \\
                      \hline
 [layer $10$] & \makecell{r\_p=$1$; Conv2d $(3,3,256)$, s=$1$; InstanceNorm; ReLU;\\
                      r\_p=$1$; Conv2d $(3,3,256)$, s=$1$; InstanceNorm;} \\
                      \hline
 [layer $11$] & \makecell{r\_p=$1$; Conv2d $(3,3,256)$, s=$1$; InstanceNorm; ReLU;\\
                      r\_p=$1$; Conv2d $(3,3,256)$, s=$1$; InstanceNorm;} \\
                      \hline
 [layer $12$] & \makecell{r\_p=$1$; Conv2d $(3,3,256)$, s=$1$; InstanceNorm; ReLU;\\
                      r\_p=$1$; Conv2d $(3,3,256)$, s=$1$; InstanceNorm;} \\
                      \hline
 [layer $13$]& DeConv2d $(3,3,128)$, s=$2$, p=$1$; InstanceNorm; ReLU;\\
 \hline
 [layer $14$]& DeConv2d $(3,3,64)$, s=$2$, p=$1$; InstanceNorm; ReLU;\\
\hline
 [layer $15$]& r\_p=$3$; Conv2d $(7,7)$, s=$1$; Tanh;\\
\hline
 \hline
 Output:& Image ($256$x$256$)\\
 \hline

 \end{tabular}
 }
 \end{center}
 \end{table}
 
 \subsubsection{Generator Architecture}
The Generator network, as shown in Table\ref{Generator Network} 
consists of $3$ Convolutional Layers followed by $9$ residual blocks and $3$ Deconvolutional Layers, is basically adapted from \cite{PLRTSTSR}. 
Instead of batch normalization we used instance normalization because in image generation networks the later one have shown great superiority over the previous one \cite{CyclicGAN}. The input image of dimension $256 \times 256$ in source domain is given to the network. The network follows a series of down convolutions and up convolutions to retain the  $256$x$256$ in another domain.

  \begin{table}[t]
 \caption{Discriminator Network Architecture. The $s$ and $p$ denote the stride and the padding, respectively.}
 \label{Discriminator Network}
 \begin{center}
 \scalebox{1}{
 \begin{tabular}{l l}
 \hline Input: &Image($256$x$256$) \\
 \hline
 \hline
 [layer $1$]& Conv2d $(4,4,64)$, s=$2$, p=$1; $ LReLU;\\
 \hline
 [layer $2$]& Conv2d $(4,4,128)$, s=$2$, p=$1$; InstanceNorm; LReLU;\\
 \hline
 [layer $3$]& Conv2d $(4,4,256)$, s=$2$, p=$1$; InstanceNorm; LReLU;\\
 \hline
 [layer $4$]& Conv2d $(4,4,512)$, s=$1$, p=$1$; InstanceNorm; LReLU;\\
 \hline
 [layer $5$]& Conv2d $(4,4,1)$, s=$1$, p=1;\\
 \hline
 \hline
 Output:& (Real/Fake) Score\\
 \hline
 \end{tabular}
 }
 \end{center}
 \end{table}

\subsubsection{Discriminator Architecture}
The used Discriminator network in the proposed method is a $70 \times 70$ PatchGAN taken from \cite{cGAN}. This network consists of the $4$ Convolutional Layers, each one is a sequence of Convolution-InstanceNorm-LeakyReLU, followed by $1$ Convolutional Layer to produce a $1$ dimensional output as shown in Table \ref{Discriminator Network}
In the $1^{st}$ convolutional layer, we do not use any normalization. 

The Discriminator network takes an image of dimension $256 \times 256$ and generates a probability of an image being either fake or real (i.e., a score between $0$ for fake and $1$ for real). The Leaky ReLUs with slope 0.2 is used as the activation function in the Discriminator network.

\section{Experimental Setup}
This section, is devoted to present the experimental setup such as datasets, evaluation metric, training details and methods compared.the 
datasets used in the proposed method are described first, followed by the description of the evaluation metrics, the training details of the proposed method and the GAN methods used for the results comparison. 

\subsection{DataSets}
To appraise the efficiency of the proposed CSGAN method, we have evaluated the method on two publicly available datasets, namely CUHK and Facades. The subsection describes the datasets in brief. 
\subsubsection{CUHK Student Dataset }
The CUHK\footnote{http://mmlab.ie.cuhk.edu.hk/archive/facesketch.html} dataset consists of $188$ face image pairs of sketch and corresponding face of students \cite{chuk}. 
 The cropped version of the CUHK dataset is used in this paper. The images are resized to the dimension of $256 \times 256$ from the original dimension of $250 \times 200$.
Among $188$ images, $100$ images are used for the training and rest for the testing. 
\subsubsection{CMP Facades Dataset}
The CMP Facades\footnote{http://cmp.felk.cvut.cz/~tylecr1/facade/} dataset has $606$ image pairs of labels and corresponding facades with dimensions of $256 \times 256$. The $400$ image pairs are used for the training and remaining for the testing. 

\subsection{Evaluation Metrics}
The quantitative as well as qualitative results are computed in this paper in order to get the better understanding of the performance of the proposed method. The Structural Similarity Index (SSIM) \cite{SSIM}, Mean Square Error (MSE) and Peak Signal to Noise Ratio (PSNR) evaluation measures are adapted to report the results quantitatively. 
 These evaluation measures are very common in image-to-image transformation problem to judge the similarity between the generated image and ground truth image. 
We also used  the Learned Perceptual Image Patch Similarity (LPIPS) metric proposed by \cite{LPIPS} and used in \cite{TMI2MT}. 
 The LPIPS calculates the distance between the real and generated images by employing the more emphasis on perceptual similarity. 
The results are also depicted in the form of the produced images along with the real images for the qualitative comparison.
\begin{table}[t]
\caption{The average scores of the SSIM, MSE, PSNR and LPIPS metrics for the proposed CSGAN and latest state-of-the art methods trained on CUHK Dataset.The values in bold highlights the best values, and the  italic represents the next best.}
\label{chuk_table}
\begin{center}
\scalebox{1}{
\begin{tabular}{|c|c|c|c|c|c|}
\hline
\textbf{Methods} & \multicolumn{4}{ c |}{\textbf{Metrics}}  \\ 
\cline{2-5}
& SSIM & MSE & PSNR & LPIPS \\
 \hline
GAN & $0.5398$ & $94.8815$ & $28.3628$ & $0.157$  \\
\hline
Pix2Pix & $0.6056$ & $89.9954$ & $28.5989$ & $0.154$   \\
\hline
DualGAN & $0.6359$ & $\mathit{85.5418}$ & $\mathit{28.8351}$ & $0.132$
\\
\hline
CycleGAN & $\mathit{0.6537}$ & $89.6019$ & $28.6351$ & $0.099$ \\
\hline
PS2GAN & $0.6409$ & $86.7004$ & $28.7779$ & $\mathit{0.098}$\\
\hline
CSGAN & $\mathbf{0.6616}$ & $\mathbf{84.7971}$ & $\mathbf{28.8693}$ & $\mathbf{0.094}$ \\
\hline
\end{tabular}
}
\end{center}
\end{table}

\begin{table}[t]
\caption{The average scores of the SSIM, MSE, PSNR and LPIPS metrics for the proposed CSGAN and latest state-of-the art methods trained on FACADES Dataset.The values in bold represents the best values, and the  italic represents the next best.}
\label{Facades_table}
\begin{center}
\scalebox{1}{
\begin{tabular}{|c|c|c|c|c|}
\hline
\textbf{Methods} & \multicolumn{4}{ c |}{\textbf{Metrics}}  \\ 
\cline{2-5}
& SSIM & MSE & PSNR & LPIPS \\

 \hline
GAN & $0.1378$ & $103.8049$ & $27.9706$ & $0.252$  \\
\hline
Pix2Pix & $\mathit{0.2106}$ & $\mathbf{101.9864}$ & $\mathbf{28.0569}$ & $\mathbf{0.216}$   \\
\hline
DualGAN & $0.0324$ & $105.0175$ & $27.9187$ & $0.259$
\\
\hline
CycleGAN & $0.0678$ & $104.3104$ & $27.9489$ & $0.248$ \\
\hline
PS2GAN & $0.1764$ & $\mathit{102.4183}$ & $\mathit{28.032}$ & $0.221$\\
\hline
CSGAN & $\mathbf{0.2183}$ & $103.7751$ & $27.9715$ & $\mathit{0.22}$ \\
\hline
\end{tabular}
}
\end{center}
\end{table}

\begin{figure*}[t]
\begin{center}
\includegraphics[width=\linewidth, height=11 cm]{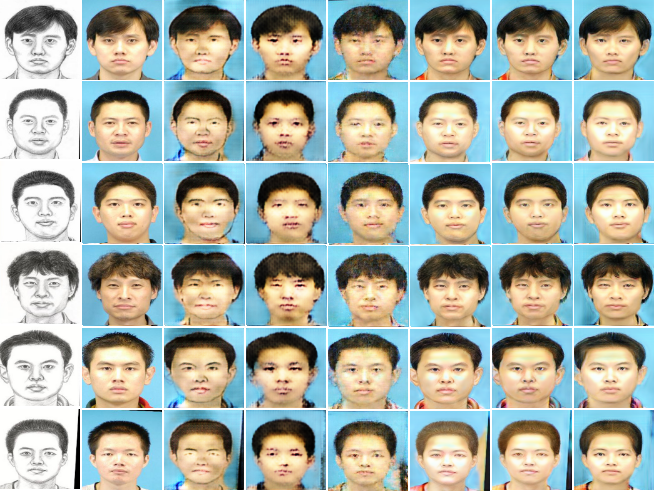}
\setlength{\belowcaptionskip}{-20pt} 
\caption{Represents qualitative resemblance of sketch to photo transformation using CUHK dataset. From left to right: Input, Ground truth, GAN, Pix2Pix, DualGAN, CyclicGAN, PS2GAN and CSGAN. The CSGAN achieves lowest artifacts and generates the realistic and fair images. }
\label{cuhk_fig}
\end{center}
\end{figure*}

\begin{figure*}[t]
\begin{center}
\includegraphics[width=\linewidth, height=11 cm]{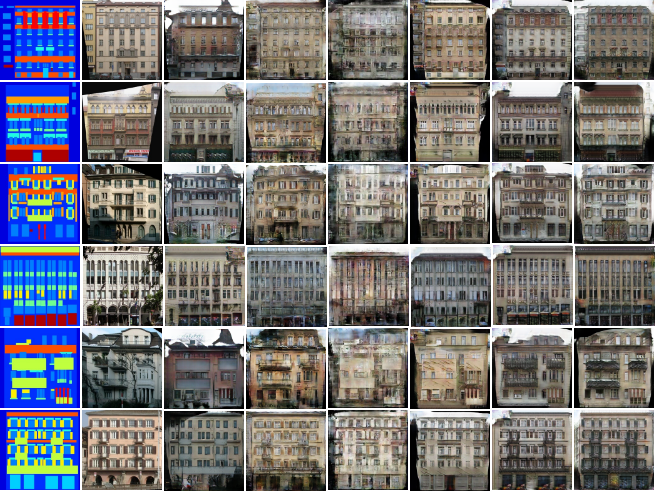}
\setlength{\belowcaptionskip}{-18pt} 
\caption{Qualitative comparison of labels to buildings transformation results on FACADES dataset. From left to right: Input, Ground truth, GAN, Pix2Pix, DualGAN, CycleGAN, PS2GAN, and CSGAN. The CSGAN generates the realistic and fair images with lowest artifacts. }
\label{facades_fig}
\end{center}
\end{figure*} 

\subsection{Training Information}
From our observation, the default settings of CycleGAN \cite{CyclicGAN} results in blurred images while trying to resize the input image from an arbitrary dimension to the fixed dimension $256 \times 256$. The resultant blurred images ultimately causes the deformations in the generated output images. So, in order to avoid this problem, the $256 \times 256$ dimensional images are used for the experiments in this paper. In each experiment, both the generator and discriminator networks are trained from scratch for $200$ epochs with the batch size as $2$.
\setlength{\belowcaptionskip}{-20pt}
The Adam solver \cite{adam} is used in this experiment for training the networks with momentum term $\beta1$ as $0.5$. It is reported in \cite{dcgan} that the higher value of $\beta1$ such as $0.9$ can lead to poor stabilization. Initially, for the first $100$ epochs, the learning rate is fix to $0.0002$ and linearly decaying to $0$ for next $100$ epochs. To initialize the network weights,we have used the Gaussian Distribution with mean as $0$ and standard deviation as $0.02$. The joint training of generator and discriminator networks are performed. 
For the proposed CSGAN, the values of the weight factors $\lambda_A$ and $\lambda_B$ both are set to $10$ and the values of the weight factors $\mu_A$ and $\mu_B$ both are set to $30$ (see Equation \ref{LFINAL}). The default settings are used for the weight factors in compared methods as per the corresponding source paper. 
Two GPUs in parallel, namely PASCAL TITAN X (12GB)
 \footnote{Generously donated by NVIDIA Corp. through the Academic Partnership Program} and GeForce GTX 1080 (8GB) are used for training the networks. 

\subsection{Compared Methods}
For analyzing the results the proposed CSGAN method is compared with five different state-of-the-art methods, namely GAN \cite{gan}, Pix2Pix \cite{cGAN}, DualGAN \cite{dualGAN}, CycleGAN \cite{CyclicGAN} and PS2MAN \cite{ps2-man}. For a fair comparison with the proposed CSGAN method as well as other methods, PS2MAN is implemented using a single adversarial network only i.e., PS2GAN. All these methods are compared with proposed method for paired image-to-image translation.

\subsubsection{GAN} The original GAN was proposed for the new sample generation from the noise vector \cite{gan}. In this paper, the Pix2Pix\footnote{https://github.com/phillipi/pix2pix \label{Pix2Pix}} \cite{cGAN} code is modified into GAN for image-to-image translation by removing the conditional property and $L_1$ loss.

 \subsubsection{Pix2Pix} The results are produced by using the code provided by the authors of Pix2Pix \cite{cGAN} with the same default settings.

 \subsubsection{DualGAN} The results are generated by using the code provided by the authors of DualGAN\footnote{https://github.com/duxingren14/DualGAN} \cite{dualGAN} with the same default settings. 

 \subsubsection{CycleGAN} The results are obtained by using the code provided by the authors of CycleGAN\footnote{https://github.com/junyanz/pytorch-CycleGAN-and-pix2pix} \cite{CyclicGAN} with the same default settings. 

 \subsubsection{PS2GAN}  The results are generated by modifying the original PS2MAN method proposed by \cite{ps2-man}, that uses a multiple adversarial networks. For a fair comparison with the proposed CSGAN method as well as other methods, PS2MAN is implemented using a single adversarial network only i.e., PS2GAN. The PS2GAN consists of the synthesized loss in addition to the losses mentioned in the CycleGAN \cite{CyclicGAN}.



\section{Experimental Results and Analysis}
In this section, results obtained by the proposed CSGAN method are compared against the five different base line image-to-image transformation methods like GAN, Pix2Pix, DualGAN, CycleGAN and PS2GAN. Both Quantitative and Qualitative analysis of the results are presented to reveal the improved performance of the proposed method. 

\subsubsection{Quantitative Evaluation}
Table \ref{chuk_table} and Table \ref{Facades_table} list the comparative results over the CUHK sketch-face and FACADES labels-buildings datasets, respectively using five different state-of-the-art methods along with the proposed CSGAN method. 
In terms of the average scores given by the SSIM, the MSE, the PSNR and the LPIPS metrics, the proposed CSGAN method clearly shows improved results over the other compared methods. It is also observed that the proposed CSGAN generates more structurally and perceptually similar faces for a given sketch as it has highest value for SSIM and lowest value for LPIPS. The lowest value of MSE and highest value of PSNR points out that the proposed method generates the faces with more pixel level similarity. 
 Thus, the proposed method is able to provide a very balanced trade-off between pixel-level similarity and structure/perceptual-level similarity.

In terms of structural similarity, i.e., the average SSIM score, the proposed CSGAN method outperforms the state-of-the-art GAN based methods. In terms of perceptual similarity, i.e., the proposed CSGAN is very close to the the best performing Pix2Pix method. One of the possible reason for it is due to the poor performance of adversarial loss itself. The performance improvement due to the proposed cyclic-synthesized loss (CS Loss) is dependent upon the performance of adversarial loss because the images used in CS Loss are not the original images, rather the synthesized and cycled images. It can be seen from the LPIPS results over FACADES dataset (see Table \ref{Facades_table}), that the LPIPS is poor for all the methods.
For the FACADES dataset and in the context of the MSE and the PSNR metrics (i.e., the pixel-level similarity), the proposed method is not able to produce the best result due to the huge amount of difference between the labels and buildings domains. 

\subsubsection{Qualitative Evaluation} 
 Fig. \ref{cuhk_fig} and Fig. \ref{facades_fig} show the qualitative comparison of the CSGAN results on CUHK and FACADES datasets, respectively, with five different state-of-the-art methods. The GAN, Pix2Pix and DualGAN are unable to generate the output even close to ground truth and contain the different type of artifacts such as face distortion, background patches, missing blocks, etc. 
 The results of PS2GAN have the brightness inconsistency as well as missing patches in different samples. 
The results of CycleGAN is reasonably better, but still suffers with the color inconsistencies for different samples.
These shortcomings are removed in the results of the proposed CSGAN method which are more realistic as compared to the other methods in terms of the shape, color, texture and reduced artifacts. 
\section{Conclusion}
In this paper, we proposed a new method for image-to-image transformation called as CSGAN. The CSGAN is based on the Cyclic-Synthesized loss. 
 Ideally, the cycled image should be similar to the synthesized image in a domain. The Cyclic-Synthesized loss finds the error between the synthesized and cycled images in both the domains.
By adding the Cyclic-Synthesized loss to the objective function (i.e., other losses such as Adversarial loss and Cycle-consistency loss), the problem of unwanted artifacts is minimized. The performance of proposed CSGAN is validated over two benchmark image-to-image translation datasets and the outcomes are analyzed with the recent state-of-the-art methods. The thorough experimental analysis, confirms that the proposed CSGAN outperforms the state-of-the-art methods. 

The performance of the proposed method is also either better or comparable over other datasets.
In future we want to extend our work towards optimizing the generator and discriminator networks and to focus on unpaired datasets i.e., towards unsupervised learning

\section*{Acknowledgement}
The authors would like to thanks the NVIDIA Corp. for donating us the NVIDIA GeForce Titan X Pascal GPU used in this research.
\bibliographystyle{IEEEtran}  
\bibliography{egbib}
\end{document}